\documentclass[letterpaper, 10 pt, conference]{ieeeconf}  
                                                       
\IEEEoverridecommandlockouts                              
\usepackage[pdftex]{graphicx}
\usepackage[cmex10]{amsmath}
\usepackage{amssymb}
\usepackage{import}
\usepackage{bm}

\usepackage{color}
\usepackage{units}
\usepackage{url}
\usepackage{float}
\usepackage{epstopdf}
\usepackage{stmaryrd}

\usepackage{units}
\usepackage{siunitx}

\usepackage{outlines}

\usepackage{boldline}
\usepackage[flushleft]{threeparttable}

\usepackage[style=ieee,backend=biber,maxbibnames=1000]{biblatex}

\bibliography{main.bib}
\usepackage{mathtools}
\usepackage{balance}

\newcommand{\mnoshow}[1]{}

\newcommand{\figref}[1]{Fig.~\ref{#1}}
\newcommand{\secref}[1]{Section~\ref{#1}}

\title{\LARGE \bf
Vehicle Localization and Control on Roads with Prior Grade Map
}
\author{Roya Firoozi, Jacopo Guanetti, Roberto Horowitz, Francesco Borrelli
\thanks{The authors are with the Department of Mechanical Engineering, at the University of California, Berkeley.
\newline        {\tt \{royafiroozi, jacopoguanetti, horowitz, fborrelli \}@berkeley.edu}}%
}

\begin{document}
\maketitle
\thispagestyle{empty}
\pagestyle{empty}
\begin{abstract}
We propose a map-aided vehicle localization method for GPS-denied environments. This approach exploits prior knowledge of the road grade map and vehicle on-board sensor measurements to accurately estimate the longitudinal position of the vehicle. Real-time localization is crucial to systems that utilize position-dependent information for planning and control. We validate the effectiveness of the localization method on a hierarchical control system. The higher level planner optimizes the vehicle velocity to minimize the energy consumption for a given route by employing traffic condition and road grade data. The lower level is a cruise control system that tracks the position-dependent optimal reference velocity. Performance of the proposed localization algorithm is evaluated using both simulations and experiments.
\end{abstract}

\section{Introduction}
Localization in autonomous driving applications refers to determining the vehicle's position and attitude. Navigation, motion planning and real-time control rely on accurate localization. Inaccurate localization can be detrimental to performance and may lead to accidents.
The Global Positioning System (GPS) is able to provide positioning with meter-level accuracy in clear open sky, but since its accuracy suffers from degraded satellite availability or multi-path error in city areas, a stand-alone GPS is not reliable. A common method to overcome these limitations is using dead-reckoning techniques. GPS data and dead-reckoning information extracted from vehicle's on-board sensors can be fused to improve the position estimate of the vehicle \cite{CARON2006221}, \cite{Sahawneh2011}. Since dead-reckoning estimates the current position based on a previously determined position, it is subject to significant cumulative error. In sparse GPS environments like tunnels, underground passages or forested paths, using only dead-reckoning may lead to large estimation errors. A solution to this problem is incorporating sensor data with a prior road map to correct the position estimates. Map-aided localization approaches exploit road features including lane markings, traffic signs, curbs and buildings to correlate the vehicle states to the prior road map.   

Today's self driving cars equipped with three-dimensional (3D) Light Detection and Ranging (LIDAR) scanners are capable of localizing themselves with centimeter-level accuracy. High accuracy localization requires a high definition map of the environment. LIDAR generates point clouds by illuminating laser pulses to the surroundings and measuring the reflected results. In order to build a high-resolution map of the environment, a survey vehicle is driven in the road and measurements of GPS, Inertial Measurement Unit (IMU), wheel odometry and LIDAR data are integrated together. Once an a-priori map of the environment is generated, the vehicle can match the features extracted from LIDAR point clouds with the features stored in the map to correct its position estimate in real-time \cite{6280129}, \cite{WANG2017276}. Since high-accuracy LIDAR is expensive, a cheaper way to do localization is to use LIDAR for mapping and camera for localization. A survey vehicle equipped with LIDAR generates a Point Cloud Map (PCL) of the road and autonomous cars can make use of just camera to localize itself within the (3D) pre-built PCL \cite{Caselitz2016MonocularCL}.

Vision-based matching of road features to a pre-built map of the environment can augment localization performance\cite{ROB20395}. However, these feature-based localization methods yield accurate localization in feature-rich environments like urban areas. In rural and suburbans areas, where density of features decreases dramatically, these approaches might not be reliable. A feature that is present in both urban and country roads is road grade. We propose to make use of a road grade map as the a-priori map for localization. The proposed localization method uses an Extended Kalman Filter (EKF) that combines data provided by the on-board sensors dead-reckoning information with the grade map to accurately estimate the vehicle's longitudinal position. Although the application of our proposed approach is restricted to hilly roads, it can correct dead-reckoning in extended GPS blackout or augment IMU/GPS system where GPS signal is available. Also this method can be combined with other techniques discussed above to improve the localization performance.  

Applications that benefit from localization include planning and control systems that rely on the route spatial data like road grade and curvature, speed limit, road optimal reference velocity trajectory and distance to the next stop sign or traffic signal. We validate our localization approach on a control architecture comprised of two levels. At the high-level, we optimize the vehicle velocity trajectory over a given route, by incorporating traffic flow and road grade data into the problem. At the low-level, we run a Model Predictive controller (MPC) that follows the optimal reference velocity trajectory calculated by high-level planner. Our simulations show how errors in position estimation affect energy consumption and confirm the effectiveness of the proposed approach.

The rest of the paper is structured as follows. \secref{secLocalization} describes road grade map generation and presents the localization approach using EKF formulation. \secref{secLocalizationRes} shows simulation and experimental results for localization. \secref{secControl} explains the hierarchical control system. \secref{secControlRes} illustrates the simulation results for control system and \secref{secConclusion} makes concluding remarks.

\section{Localization}\label{secLocalization}
In case of temporary GPS loss, odometry sensors such as wheel speed encoder can be used to estimate the change of position. However, this method is sensitive to errors due to the integration of velocity measurements over time. To mitigate this cumulative error, we propose to fuse the sensor measurements from wheel speed encoders and accelerometer with grade data extracted from a global road map using the EKF.
This section describes road grade map generation and process and measurement models that are included in the EKF estimator. 
\subsection{Road Grade Map Generation}
Roadway elevation data is crucial for many transportation applications. To generate a grade map for the entire road, a high-precision Real Time Kinematic (RTK) GPS can be used to measure latitude, longitude and altitude along the route. Also several APIs such as Google Elevation API provide elevation data for all locations on earth surface. We can query elevation data for specified coordinates and obtain the road elevation profile. This API also provides the resolution of each elevation sample, defined as the maximum horizontal distance between data points from which the elevation was interpolated. Digital Elevation Models (DEMs) data throughout the earth's surface has been provided by the NASA Shuttle Radar Topographic Mission (SRTM). This elevation data in the United States is available in National Elevation Dataset (NED), provided by the Geological Survey (USGS), at resolutions between 1 arc-second (about 30 meters) and $1/9$ arc-second (about 3 meters) depending on the location. According to \cite{10.1371/journal.pone.0175756} that assessed the accuracy of road elevation data extracted from Google Earth, root mean squared error (RMSE) and standard deviation of roadway elevation error are 2.27 meters and 2.27 meters, respectively, even in the areas where USGS NED provides $1/3$ arc-second (about 10 meters) resolution. Thus, Google Elevation API provides sufficient accuracy and resolution for generating road grade map.    
\subsection{Extended Kalman Filter (EKF)}
\begin{figure}
    \centering
    \includegraphics[scale = 0.6]{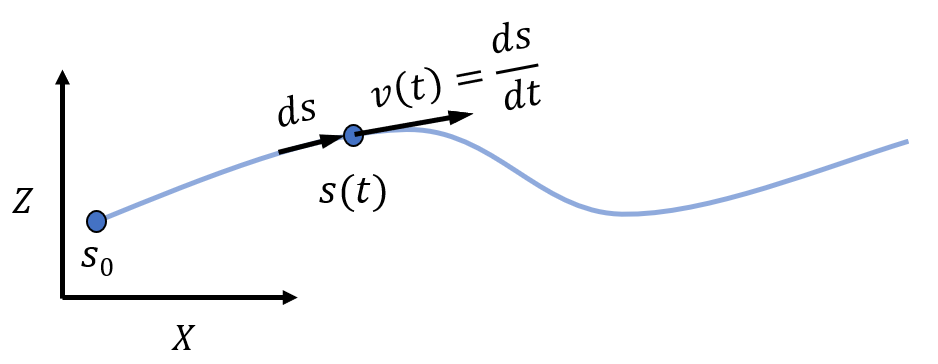}
    \caption{Point mass model of the vehicle traveling along a hilly road. The Z axis is the altitude and the X axis is the horizontal distance.}
    \label{fig:system_states}
\end{figure}

We model the vehicle as a point mass moving along a path with velocity $v$, as seen in \figref{fig:system_states}. The system state at time t is
\begin{equation*}
x(t) = [s(t) \ v(t)]^{T},
\end{equation*}
where $s(t)$ and $v(t)$ are the vehicle position and velocity respectively. The process model is a nonlinear function $f$ of the state $x$ and control input $u$. Process noise $w$ is assumed to be additive and normally distributed with zero mean and covariance of $Q$, $ w \sim \mathcal{N}(0,Q)$.
The measurement model is also a nonlinear function $h$ that maps the state $x$ to output $y$. The measurement noise is assumed to be additive Gaussian noise with zero mean and covariance of $R$, $\eta \sim \mathcal{N}(0,R)$. EKF linearizes the nonlinear system about the current estimates.
\begin{equation}\label{eq:general_ekf}
\begin{aligned}
\dot{x} &= f(x,u) + Gw \\
y &= h(x) + \eta. 
\end{aligned}
\end{equation}
\subsection{Process model}
The longitudinal kinematic model of the vehicle is defined as 
\begin{align*}
\dot{s}&= v \\
\dot{v}&= a + w, 
\end{align*}
where $a$ is the longitudinal acceleration treated as a known input. The system is discretized using Euler method with constant sampling time interval $\Delta t$. 

\begin{align*}
s(t+1)&= s(t) + v(t) \ \Delta t\\
v(t+1)&= v(t) + (a(t)  + w(t))\ \Delta t.
\end{align*}
Since $\Delta t$ is small, the second-order terms are neglected. The longitudinal acceleration $a(t)$ is derived from the acceleration measured by the vehicle accelerometer. When a vehicle is moving on an inclined road, the longitudinal acceleration of the sensor is affected by gravity, 
\begin{equation}\label{eq:accelerometer_signal}
a_{sensor}(t) = a(t) + g \sin (\theta (t)),
\end{equation}
where $a_{sensor}$ denotes acceleration measured by the accelerometer, $g$ is gravitational acceleration and $\theta$ is slope of the road. 
By substituting $\sin (\theta)$ with a prior known grade map $p(s)$, longitudinal acceleration is calculated as 
\begin{equation*}
a(t) = a_{sensor}(t) - g \ p(s(t)).
\end{equation*}
Thus, the prediction model is
\begin{equation}\label{eq:discretized_process_model}
\begin{aligned}
s(t+1) &= s(t) + v(t) \ \Delta t\\
v(t+1)  &= v(t) + (a_{sensor}(t) - g \ p(s(t))) \Delta t  + w(t) \ \Delta t.
\end{aligned}
\end{equation}
\subsection{Measurement Model}
The vehicle speed $v_{m}$ is measured by the wheel speed encoders and the inclination $\theta(t)$ can be constructed by measurements of IMU accelerometer and wheel speed encoder. By reformulating \eqref{eq:accelerometer_signal} the inclination $\theta(t)$ can be indirectly measured as 
\begin{equation}\label{eq:measured_theta}
 \theta_{m} = \sin^{-1}\left(\frac{a_{sensor} - \dot{v}}{g}\right),
\end{equation}
where $\dot{v}$ is the vehicle acceleration obtained by differentiating the longitudinal velocity measured from the wheel speed sensor. Measurement noises for both IMU and wheel encoder sensors are assumed to be additive with Gaussian distributions $\eta^{v} \sim \mathcal{N}(0,\sigma_{v}^2)$ and $\eta^{\theta} \sim \mathcal{N}(0,\sigma_{\theta}^2)$. The measurement covariance matrix is defined as $ R = \ diag (\sigma_{v}^2,\sigma_{\theta}^2)$. The measurement model is 
\begin{equation}\label{eq:measurement_model}
\begin{aligned}
v_{m}(t) &= v(t) + \eta^{v}\\
\theta_{m}(t) &= \sin ^{-1}(p(s(t))) + \eta^{\theta}.
\end{aligned}
\end{equation}
\eqref{eq:discretized_process_model} and \eqref{eq:measurement_model} define the $f$ and $h$ functions in the general formulation \eqref{eq:general_ekf}, respectively.
\subsection{Inertial Sensor Noise Filtering and Bias Removal}
Since the acceleration signal measured by IMU is noisy, a low pass filter is used for noise reduction. Also, inertial sensors are subject to bias drift. In the absence of input, the IMU accelerometer reads non-zero output. This bias should be compensated, but the challenge is that bias drifts over time due to temperature changes and other factors. Hence, bias drift should be modeled mathematically and removed from the measurements. In this study we assume the bias drifts linearly as a function of time. Therefore \eqref{eq:measured_theta} is reformulated as 
\begin{equation}\label{eq:measured_theta_with_bias}
 \theta_{m} = \sin^{-1}(\frac{a_{sensor} - \dot{v}}{g}) - \underbrace{bt}_\text{bias},
\end{equation}
where $t$ is the time and $b$ is the bias parameter. We will discuss bias parameter estimation in more detail in the next section.

\section{Localization Results}\label{secLocalizationRes}
\begin{figure}
\centering
\includegraphics[scale = 0.21]{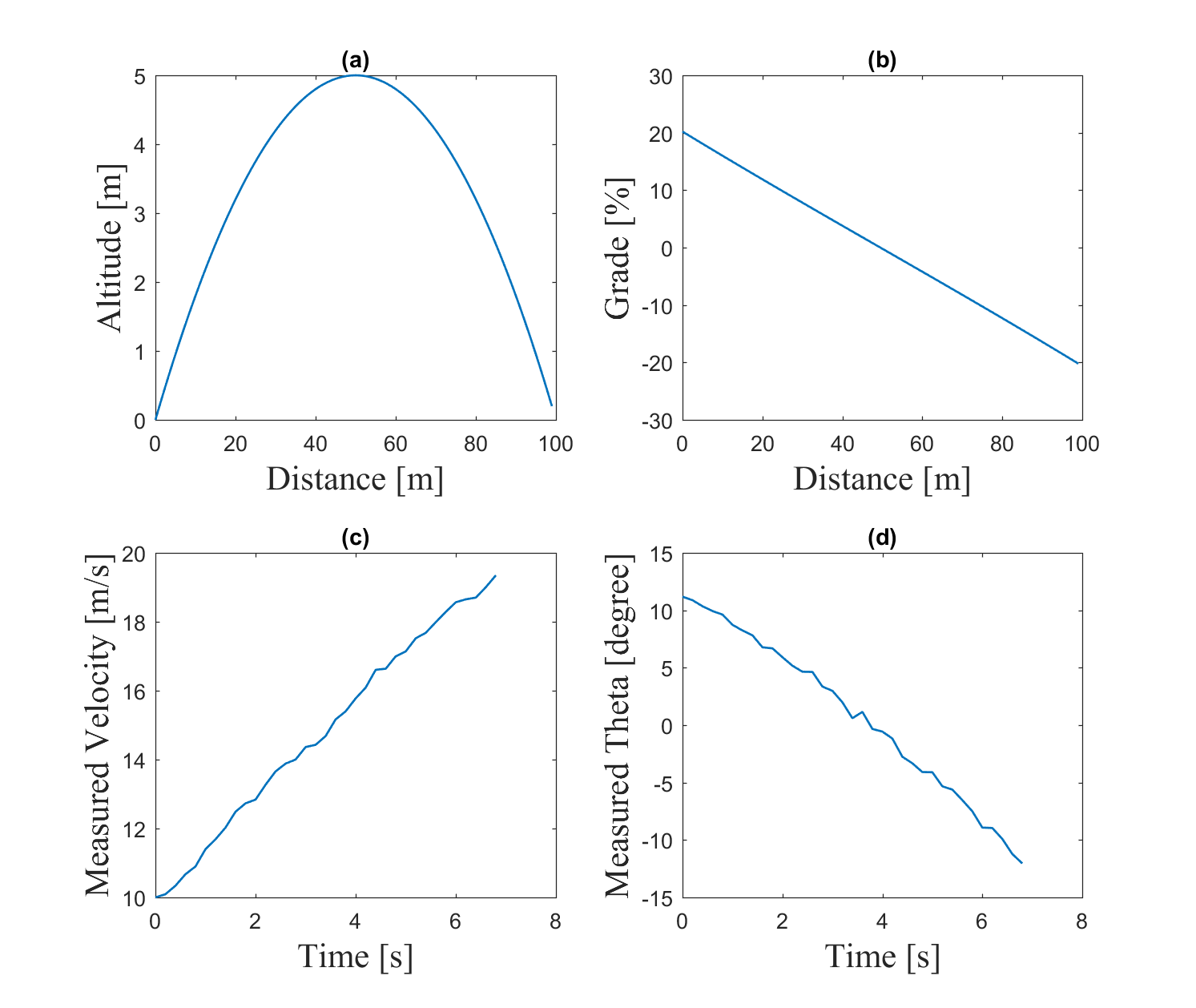}
\caption{Altitude and grade maps as well as velocity and inclination signals are generated to be used by simulator.}
\label{fig:generated_signals}
\end{figure}

To validate the effectiveness of the proposed localization approach, we test the algorithm with both simulations and experiments. Assuming that the GPS signal is lost, we compare the localization performance of our method to velocity integration. In simulation, the road altitude map is assumed to be a polynomial function of position as depicted in \figref{fig:generated_signals}a. Measured velocity and inclination signals are generated by assuming Gaussian noise. \figref{fig:generated_signals} shows the generated maps and signals. The absolute position error $e$ is defined as

\begin{equation*}
e = \hat{s}-s,
\end{equation*}
where $\hat{s}$ is the estimated position and $s$ is the true position. In Table \ref{tab:rmse}, root-mean-square error (RMSE) values for ten runs of simulation are displayed and their average values are calculated. As seen, the average RMSE using the EKF estimation approach is considerably smaller than using velocity integration. Absolute error between the estimate from velocity integration and the ground truth increases over time while the absolute error between Kalman estimate and ground truth is non-increasing and bounded, according to the simulation results.     

\begin{figure}
\centering
\includegraphics[scale = 0.21]{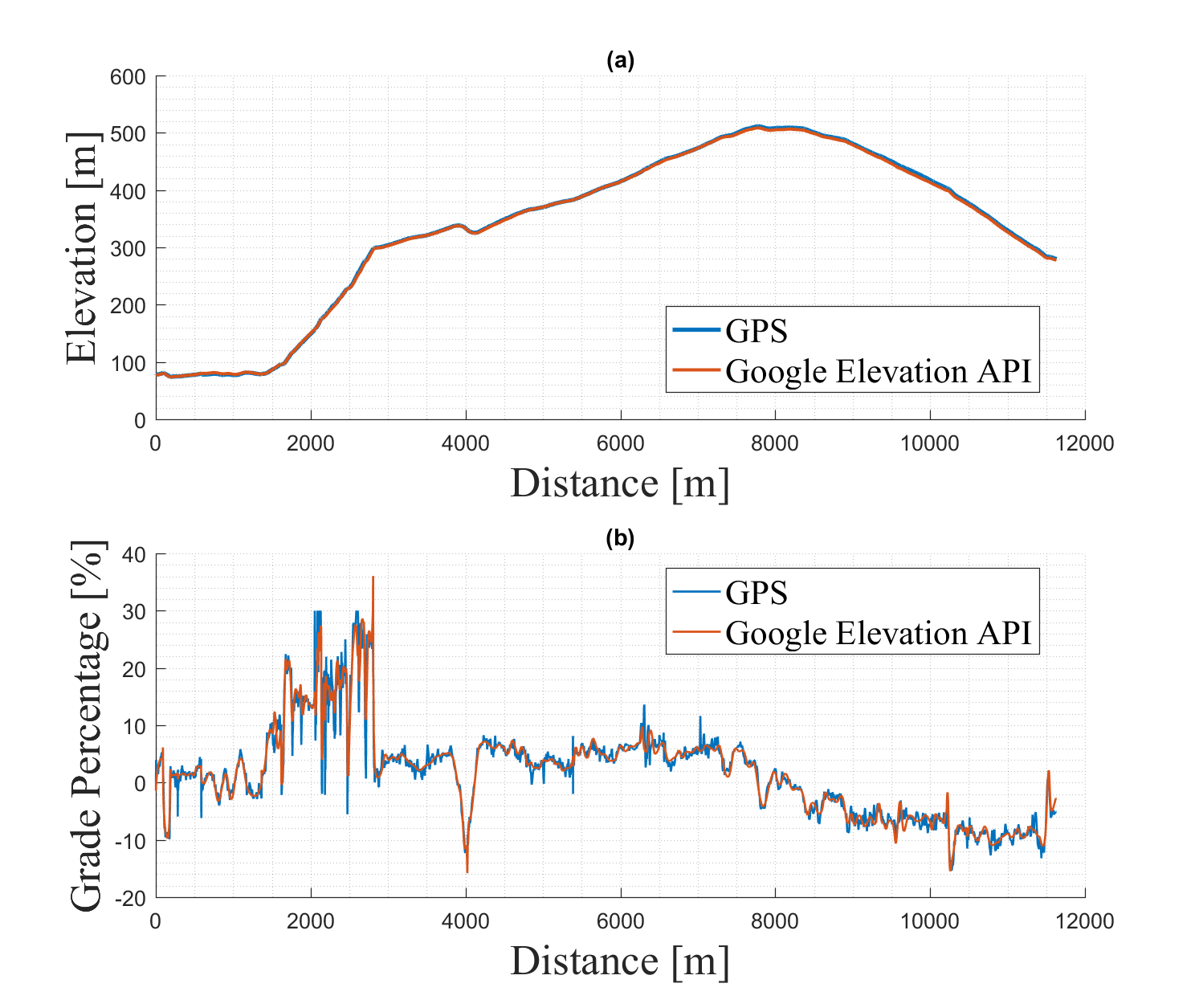}
\caption{Elevation and grade maps are generated using both GPS and Google Elevation API for a hilly route located near UC Berkeley.}
\label{fig:Altigrademap}
\end{figure}

\begin{table}
\centering
\caption{Simulation results: comparison of root-mean-square error}
\begin{tabular}{| c | c | c |}
\hline
\multicolumn{3}{|c|}{Position RMSE [m]}\\
\hline
Iteration & Velocity Integration & EKF Estimate 
\\
\hline \hline
1 &1.04 & 0.12 \\
2 &1.20 & 0.19\\
3 &0.50 & 0.16 \\
4 &0.64 & 0.12 \\
5 &0.96 & 0.12 \\
6 &0.70 & 0.09\\
7 &0.74 & 0.20\\
8 &0.87 & 0.14\\
9 &0.83 & 0.16 \\
10 &0.81 & 0.14 \\
\hline
Average & 0.83 & 0.14 \\
\hline
\end{tabular}
\label{tab:rmse}
\end{table}
The performance of the algorithm has also been examined through an experiment. The experiment was carried out in a hilly road located near University of California Berkeley. The test vehicle is a Hyundai Genesis equipped with OTS (Oxford Technical Solutions) RT2002 sensing system which is comprised of a high precision GPS and an IMU. RT2002 GPS without its base-station GPS receiver provides positioning with 0.6m Circular Error Probable (CEP) accuracy. We drove the car on the road and collected the position and altitude data measured by the RT2002 GPS system to build a high-resolution road grade map. The obtained map is compared with the map generated by Google elevation API in \figref{fig:Altigrademap}. 
To measure the inclination, bias drift is modeled as a linear function of time and removed from the measurement. To determine the mathematical model of the acceleration bias drift, first the road slope profile for a segment of the road is computed using \eqref{eq:measured_theta} with collected acceleration and velocity data. Then the computed slope profile $\theta_{n}$ is compared against the slope profile  obtained by Google Elevation API $\theta_{Google}$. The difference between the two profiles is equivalent to the bias term in \eqref{eq:measured_theta_with_bias},
\begin{equation*}
e_{slope}(t) = \theta_{n}(t) - \theta_{Google}(t) \simeq bt.
\end{equation*}
The bias parameter $b$ is estimated by fitting a linear function to the above slope error profile $e_{slope}(t)$ using least squares. In this study, the high-precision RT2002 IMU is used to obtain the acceleration measurements. However, by accurate modeling of the acceleration bias drift, the vehicle's on-board accelerometer can be used for acceleration measurement.    
After generating the high-resolution grade map of the road, the position is estimated using the proposed localization approach. Although we don't use GPS data to estimate the position, we still collect them and use them as the ground truth. We ran 3 experiments and calculated RMSE for both methods in Table \ref{tab:car_rmse}. As shown, the EKF algorithm achieves significantly smaller RMSE than velocity integration.   

\begin{table}
\centering
\caption{Experiment results: comparison of root-mean-square error}
\begin{tabular}{| c | c | c |}
\hline
\multicolumn{3}{|c|}{Position RMSE [m]}\\
\hline
Experiment & Velocity Integration & EKF Estimate 
\\
\hline \hline
1 &25.0 & 3.7 \\
2 &17.9 & 9.4 \\
3 &21.3 & 4.3 \\
\hline
Average &21.4  &5.8  \\
\hline
\end{tabular}
\label{tab:car_rmse}
\end{table}

Table \ref{tab:final_position} compares the drift error of the two methods for just the first experiment. As seen, the estimated position calculated by velocity integration drifts over time. The absolute error between the velocity integration estimated position and the ground truth at final point is 30 times larger than EKF estimated position. The experiment results are in agreement with simulation results and EKF estimates are relatively matched with the truth, whereas position estimates calculated by integration of wheel speed diverge as time goes on. 
\begin{table}
\centering
\caption{Experiment Results: Comparison of Drift Error of Final Position Estimate}
\begin{tabular}{| c | c | c |}
\hline
Method & Absolute Error [m] & Error Percentage [\%]\\
\hline \hline
Velocity Integration & 60.3 & 0.52 \\
\hline
EKF estimate & 2.4 & 0.02\\
\hline
\end{tabular}
\label{tab:final_position}
\end{table}

\section{Control System Architecture}\label{secControl}
\begin{figure}
\centering
\includegraphics[scale=0.6]{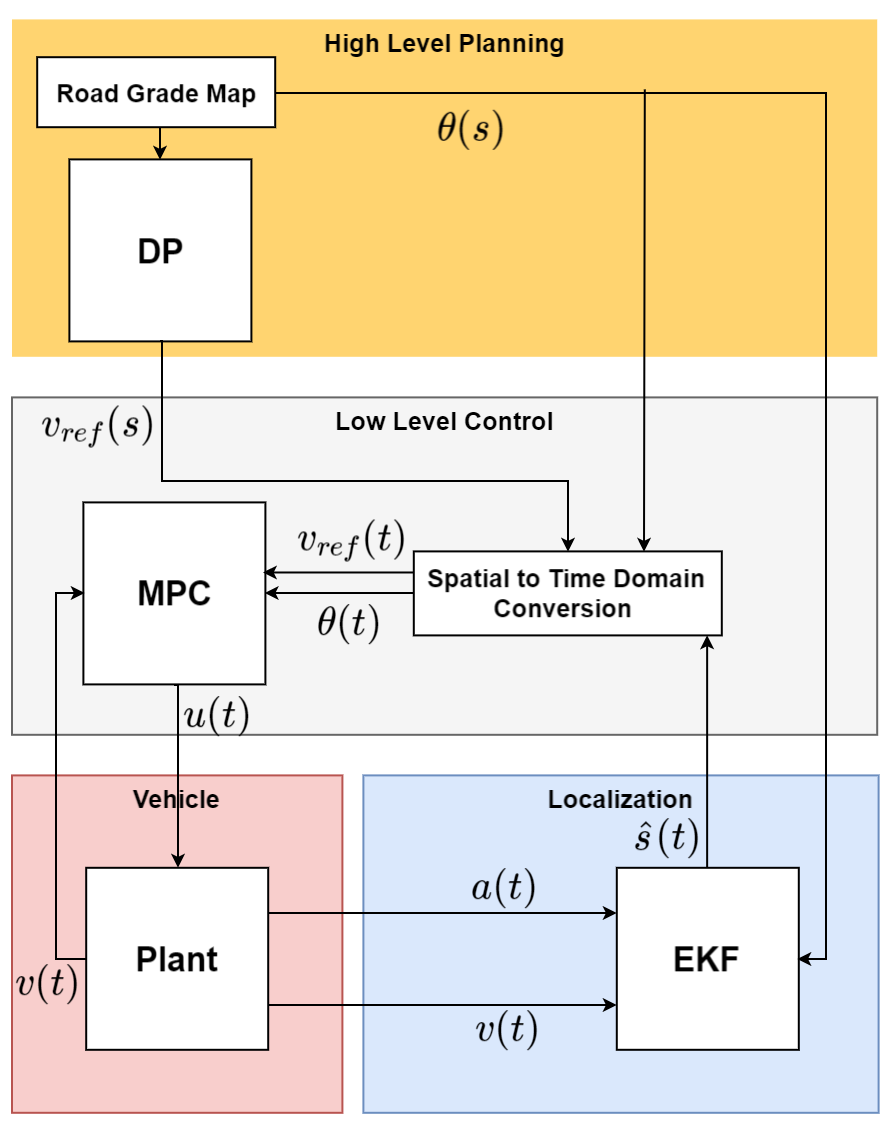}
\label{fig:archi}
\caption{System Architecture: position-dependent data including reference velocity $v_{ref}(s)$ and road grade $\theta(s)$ are converted from spatial domain to time domain}
\end{figure}

In this section we describe an example application of the proposed localization algorithm, a hierarchical control system that exploits position-dependent information for planning and control. The system architecture is shown in \figref{fig:archi}. The goal is to plan an energy-optimal velocity trajectory for a given route offline and to follow the obtained reference trajectory in real-time. Both the offline and online formulations with the vehicle longitudinal model are explained in the following sections. 

\subsection{Vehicle Model}
The longitudinal motion of a vehicle moving on an inclined road can be modeled by the following force balance in the longitudinal direction
\begin{equation}\label{eq:longitudinal_dynamics}
m\dot{v}= \underbrace{F_{traction}-F_{brake}}_\text{u(t)}-F_{airdrag}-F_{rolling}-F_{gravity},
\end{equation}
where $m$ is mass of the vehicle, $\dot{v}$ is the longitudinal acceleration and $F_{traction}$ and $F_{brake}$ are throttle and brake forces, respectively and $F_{traction} - F_{brake}$ is the total force taken as control input $u(t)$. The aerodynamic drag is determined by vehicle speed $v$, air density $\rho$, air drag coefficient $C_d$ and frontal area $A_f$.
\begin{equation*}
F_{airdrag} = \frac{1}{2}\rho C_{d}A_{f} v^2.
\end{equation*}
The rolling resistance is defined as 
\begin{equation*}
F_{rolling} = m g C_{r} \cos (\theta),
\end{equation*}
where $g$ is gravitational force, $C_{r}$ is rolling friction coefficient and $\theta$ is road slope.  
The gravity force due to road grade can be expressed as
\begin{equation*}
F_{gravity} = m g \sin (\theta).
\end{equation*}

\subsection{High-Level Planner: Dynamic Programming Formulation}
Traffic condition and roadway grade have a significant impact on the vehicle's energy consumption and emission. For a given route with specified origin and destination, the energy-optimal velocity trajectory along the route can be determined by employing information about traffic condition and road grade. To find the optimal position-dependent velocity profile, it is convenient to convert the optimization problem formulation from time domain to spatial domain,
\begin{equation*}
\dot{v} = \frac{dv}{dt} = \frac{dv}{ds}\frac{ds}{dt} = \frac{dv}{ds}v.
\end{equation*}
Velocity $(v)$ is a state variable and travelled distance $(s)$ is the independent variable. The vehicle longitudinal dynamics \eqref{eq:longitudinal_dynamics} in the spatial domain is
\begin{align}
\label{eq:spatial_dynamics}
\frac{dv}{ds} &= \frac{1}{mv}\underbrace{(u(s)-F_{airdrag}-F_{rolling}-F_{gravity})}_{\text{$F_{total}$}}
\end{align}
and after Euler discretization the dynamics is reformulated as
\begin{equation*}
v(s+1) = v(s)
+\frac{\Delta s}{mv(s)}F_{total},
\end{equation*}
where $F_{total}$ is the total longitudinal force. This formulation is singular when the velocity is zero. To avoid the singularity, by assuming a constant acceleration over the integration interval $ds$, the longitudinal dynamics \eqref{eq:spatial_dynamics} is discretized as
\begin{equation}
v(s+1)=  \\\label{eq:spatial_dynamics_nonsingular}
\big(v(s)^2+\frac{2\Delta s}{m}F_{total}\big)^\frac{1}{2},
\end{equation}
with trapezoid discretization instead of Euler method.  

The goal is to minimize the total power consumption over the entire route within a reasonable trip time interval. Therefore, the cost function is a trade-off between power consumption and trip time. The first term of the cost function \eqref{eq:planner_cost} penalizes the wheel power consumption and the second term compensates deviation from maximum speed profile to guarantee a short trip time. The objective
\begin{equation}\label{eq:planner_cost}
J(s_{cur}) = \sum_{s=s_{cur}}^{s_{final}} \big(v(s) u_{p}(s) + \gamma (v(s) - v_{max}(s))^{2}\big) \Delta s
\end{equation}
is minimized, where $s_{cur}$ and $s_{final}$ represent the current and final positions respectively. $\gamma$ is a weight factor, $u_{p}(s)$ denotes acceleration force in longitudinal direction and $v_{max}(s)$ is velocity upper bounds at position $s$. Although $u(s)$ can hold positive or negative values, $u_{p}(s)$ in the cost function \eqref{eq:planner_cost} refers to only positive inputs. The problem is formulated as
\begin{equation}\label{eq:dp_formulation}
    \begin{aligned}
    & \underset{{u(s)}}{\text{min}}
    & J &= \sum_{s=s_{cur}}^{s_{final}}\big(v(s) u_{p}(s) \\
    & & &+ \gamma (v(s) - v_{max}(s))^{2}\big) \Delta s\\
    &\textrm{s.t.} & & v(s+1) = f(v(s),u(s)),\\
    & & & v_{min}(s) \leq v(s) \leq v_{max}(s),\\
    & & & u_{min} \leq u(s) \leq u_{max},\\
    & & & u_{p}(s) = u(s) \quad\text{if} \quad u(s) > 0\\
    & & & u_{p}(s) = 0 \quad\text{if} \quad u(s) \leq 0\\ 
    & & & v(0) = 0,\\
    & & & v(s_{final}) = 0,
    \end{aligned}
\end{equation}
where the objective \eqref{eq:planner_cost} is minimized while satisfying the longitudinal dynamics \eqref{eq:spatial_dynamics}, denoted as $f$, as well as state and input constraints and boundary conditions. $v_{min}(s)$ and  $v_{max}(s)$ are velocity lower and upper bounds at position $s$ and $u_{min}$, $u_{max}$ are control inputs lower and upper bounds, respectively. The velocities at the origin and destination are assumed to be zero.   

The nonlinear optimization problem \eqref{eq:dp_formulation} is solved by dynamic programming (DP) with the solver from \cite{Sundstrm2009AGD}. The state and input spaces are discretized according to their corresponding upper and lower bounds. To incorporate the effect of traffic condition, the velocity at each step is upper bounded according to the traffic flow data acquired for a specific route. If data is not available, the road speed limit $v_{max}(s)$ is used as upper bound. Road slope information of the route is also included in the vehicle dynamic model \eqref{eq:spatial_dynamics_nonsingular}. Thus, the DP algorithm makes use of the road geometry data to find the optimal reference velocity trajectory. Starting from destination, the DP proceeds backward and evaluates the optimal cost-to-go function \eqref{eq:planner_cost} at each node based on Bellman's principle of optimality.
The backward sweep outputs a map of optimal cost and optimal policy over the state grid. In the forward sweep, DP starts from the route origin and applies the obtained optimal control policy to calculate the optimal velocity trajectory. The optimized velocity $v(s)$ will be set as the reference velocity $v_{ref}(s)$ for the low-level controller.
\subsection{Low-Level Controller: MPC Formulation}
The low-level controller is a cruise control system that controls the longitudinal dynamics of the vehicle to track the reference velocity generated by the high-level planner and minimize the control effort in real time. To design the controller, we implemented a Model Predictive Control (MPC) algorithm. The proposed controller solves the constrained finite horizon optimization problem
\begin{equation}\label{eq:MPC_formulation}
    \begin{aligned}
        & \underset{{u(.|t)}}{\text{min}}
        & J &= \sum_{k=t}^{t+N}||v(k|t)-v_{ref}|| \\
        & & &+ \gamma \sum_{k=t}^{t+N-1}||u(k|t)|| + p(v(t+N|t))\\
        &\textrm{s.t.} & & v(k+1|t) = f(v(k|t),u(k|t)),\\
        & & & v_{min} \leq v(k|t) \leq v_{max},\\
        & & & u_{min} \leq u(k|t) \leq u_{max},\\
    \end{aligned} 
\end{equation}
where $N$ is the MPC horizon, $v(k|t)$ and $u(k|t)$ are the state variable and control input at step k predicted at time t, respectively, $\gamma$ is a weight factor, $f$ represents the vehicle longitudinal dynamics \eqref{eq:longitudinal_dynamics} and $v_{ref}$ is the reference velocity trajectory obtained from the high level planner. A terminal cost $p$ is introduced as a constant large value. 

The proposed MPC controller extracts grade data from the provided road grade map. It incorporates this prior knowledge to accurately predict the future longitudinal velocity of the vehicle over the horizon and plan accordingly. Both grade and optimal reference velocity are defined as position-dependent profiles for a specific route. In order to employ these data, vehicle has to localize itself and estimate its position with respect to the map. Since our MPC operates in the time domain and route information is available in the spatial domain, these forecasts are projected in the time domain assuming constant velocity over the horizon of MPC. 

\section{Control Results}\label{secControlRes}
\subsection{High-level planner}
An urban route near UC Berkeley (from UC Berkeley Etcheverry Hall to the Richmond Field Station) is selected and traffic flow velocity as well as grade map are obtained through the HERE API and Google Elevation API, respectively. The traffic flow velocity profile is taken as the upper-bound for velocity. The lower-bound is taken as half of the upper bound. The parameters of the longitudinal dynamic model are shown in Table \ref{tab:model_parameters}. The controller parameters are presented in Table \ref{tab:controller_parameters}.
\begin{table}
\centering
\caption{Model Parameters}
\begin{tabular}{ c  c  c  c }
\hline
m & vehicle mass & kg & 1360\\
$A_{f}$ & vehicle frontal surface& $m^{2}$ & 2.30\\
$\rho$ & air density & $kg/m^{3}$ & 1.225 \\
$C_{d}$ & vehicle drag coefficient &- & 0.24\\
$C_{r}$ & vehicle roll coefficient& -& 0.01 \\
$T_{s}$ &sampling time  &s & 0.2  \\
\hline
\end{tabular}
\label{tab:model_parameters}
\end{table} 
\begin{table}[ht]
\centering
\caption{Controller Parameters}
\begin{tabular}{ c  c  c  c }
\hline
$v_{min}$ & minimum velocity & $m/s$ & traffic speed lower bound\\
$v_{max}$ & maximum velocity & $m/s$ & traffic speed upper bound\\
$u_{min}$ & minimum control input & $kN$ & -3 \\
$u_{max}$ & maximum control input & $kN$ & 3 \\
\hline
\end{tabular}
\label{tab:controller_parameters}
\end{table}
\begin{figure}
\centering
\includegraphics[scale = 0.21]{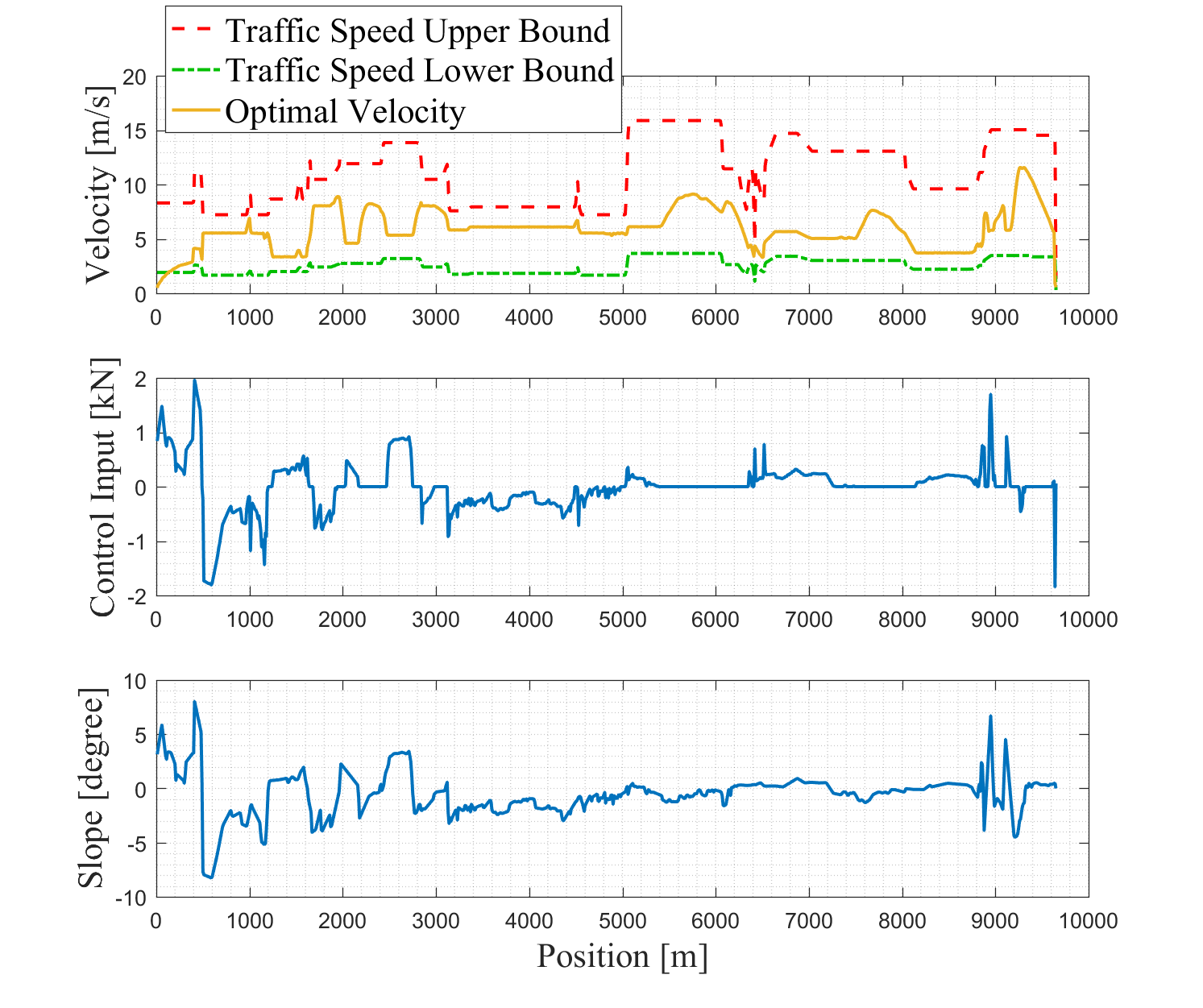}
\caption{Optimal velocity is bounded by traffic speed profile upper and lower bounds and control policy is optimized based on road slope map.}
\label{fig:DPResults}
\end{figure}
\figref{fig:DPResults} illustrates the traffic flow velocity profile as well as the velocity profile obtained by trajectory planning optimization carried out via DP. The grade map of the road is also illustrated. Table \ref{tab:DPResults} presents energy consumption as well as total trip time for traffic flow velocity profile and optimal velocity profile. As seen, the energy consumption with the optimal velocity profile (for $\gamma = 10$) is $47.8\%$ less than with the traffic flow profile, while the trip time is $26.0\%$ longer. 
\begin{table}
\centering
\caption{DP Results: comparison of the two velocity profiles}
\begin{tabular}{| c | c | c |}
\hline
Velocity Profile & Energy Consumption & Trip Time\\
&[kWh] & [min]
\\
\hline \hline
Traffic Flow & 0.9 & 20.4 \\
Optimal $(\gamma=10)$ & 0.47 & 25.7 \\
Optimal $(\gamma=0.1)$ & 0.4 & 31.4 \\
\hline
Improvement [\%]$(\gamma=10)$ & 47.8 \%  & -26.0 \%  \\
Improvement [\%] $(\gamma=0.1)$ & 55.5 \%  & -53.9 \%  \\
\hline
\end{tabular}
\label{tab:DPResults}
\end{table}

\subsection{Low-Level Controller}

\begin{figure}
\centering
\includegraphics[scale = 0.21]{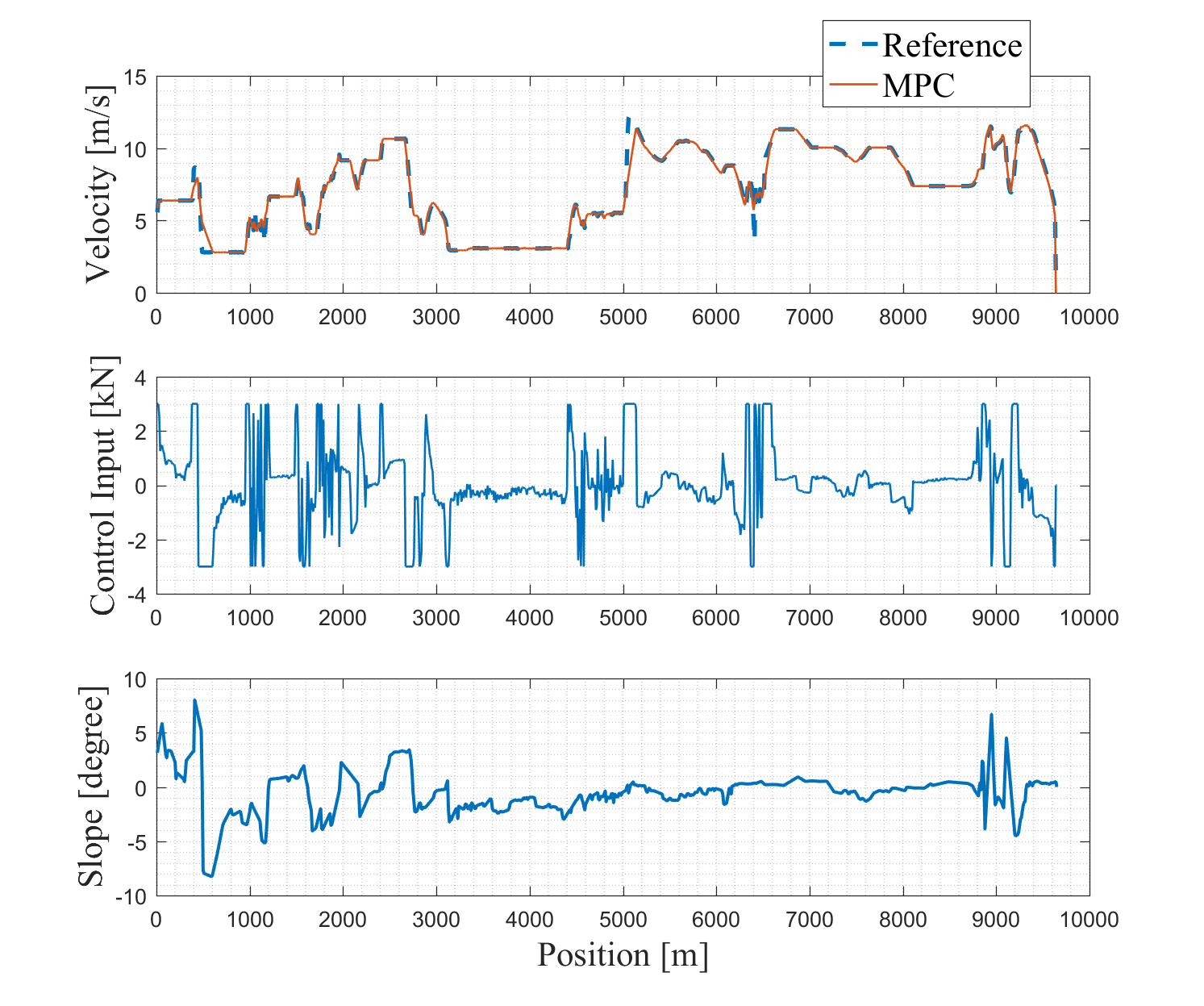}
\caption{MPC controller tracks the optimal reference velocity trajectory obtained from high-level planner and simultaneously minimizes the control input by incorporating road grade data in its short-term planning.}
\label{fig:MPCResults}
\end{figure}

We formulated the optimization problem \eqref{eq:MPC_formulation} in YALMIP \cite{YALMIP}. MPC parameters are the same as DP parameters presented in Table \ref{tab:controller_parameters} and the MPC horizon is selected as 5. \figref{fig:MPCResults} shows the simulation results of the MPC controller in closed-loop with the vehicle longitudinal dynamics model. As illustrated, the controller tracks the optimal reference velocity calculated in previous section. At the same time it minimizes the control effort by employing the road prior grade knowledge.      

Now by assuming $0.5\%$ of localization error based on the results in the table \ref{tab:final_position}, after about less than $12$ km of traveling, the position estimate can be off by $60$ meters. If the controller extracts position-dependent data including optimal reference velocity and road grade associated with incorrect position, the energy efficiency can significantly be affected by this error. As an example, we selected a portion of the road shown in \figref{fig:MPCResults} between position $8500$ m to $9500$ m and assumed that  the controller employs knowledge of $60$ meters ahead instead of its true position.
Table \ref{tab:localizationError} compares the energy consumption for both cases. The results show that energy efficiency can be affected by about $15\%$ in case of a $0.5\%$ localization error.   
\begin{table}
\centering
\caption{Energy Consumption Comparison}
\begin{tabular}{| c | c |}
\hline
Position & Energy Consumption [kWh] 
\\
\hline \hline
True Position & 0.20 \\
60 meters ahead & 0.23 \\
\hline
\end{tabular}
\label{tab:localizationError}
\end{table}

\section{Conclusion}\label{secConclusion}
We presented a localization method for autonomous driving in GPS-denied areas using a prior grade map. According to the results when there is no GPS signal, vehicle can use its own on-board sensors and a prior road grade map to localize itself relative to the map and correct its position estimate. furthermore, we developed an energy-efficient control system by exploiting traffic information as well as road slope and speed limit data. We verified that errors in localization can significantly impact energy efficiency.


\printbibliography 

\end{document}